\documentclass[letterpaper, 10 pt, conference]{ieeeconf}

\IEEEoverridecommandlockouts

\usepackage{graphicx}
\usepackage{subcaption}
\usepackage[hidelinks]{hyperref}
\usepackage{amsmath}

\usepackage{listings}
\usepackage{color}
\usepackage{booktabs}
\usepackage{multirow}
\usepackage{makecell}
\usepackage[frozencache,cachedir=.]{minted}
\usepackage{alltt}
\usepackage{hyperref}

\title{\LARGE \bf
Structured Interfaces for Automated Reasoning with 3D Scene Graphs
}

\author{Aaron Ray$^{*}$, Jacob Arkin$^{*}$, Harel Biggie, Chuchu Fan, Luca Carlone, Nicholas Roy\\\\
{*} Denotes equal contribution\\
Massachusetts Institute of Technology,\\
77 Massachusetts Avenue, Cambridge, MA 02139.\\\texttt{\{aaronray, jarkin, harelb, chuchu, lcarlone, nickroy\}@mit.edu}}

\begin{document}

\maketitle
\thispagestyle{plain}
\pagestyle{plain}

\begin{abstract}

In order to provide a robot with the ability to understand and react to a user's natural language inputs, the natural language must be connected to the robot's underlying representations of the world.
Recently, large language models (LLMs) and 3D scene graphs (3DSGs) have become a popular choice for grounding natural language and representing the world.
In this work, we address the challenge of using LLMs with 3DSGs to ground natural language.
Existing methods encode the scene graph as serialized text within the LLM's context window, but this encoding does not scale to large or rich 3DSGs.
Instead, we propose to use a form of Retrieval Augmented Generation to select a subset of the 3DSG relevant to the task. We encode a 3DSG in a graph database and provide a query language interface (Cypher) as a tool to the LLM with which it can retrieve relevant data for language grounding.
We evaluate our approach on instruction following and scene question-answering tasks and compare against baseline context window and code generation methods.
Our results show that using Cypher as an interface to 3D scene graphs scales significantly better to large, rich graphs on both local and cloud-based models. This leads to large performance improvements in grounded language tasks while also substantially reducing the token count of the scene graph content. A video supplement is available at \url{https://www.youtube.com/watch?v=zY_YI9giZSA}.

\end{abstract}

\section{Introduction} \label{sec:introduction}

\begin{figure*}[t]
    \centering
    \includegraphics[width=0.90\linewidth]{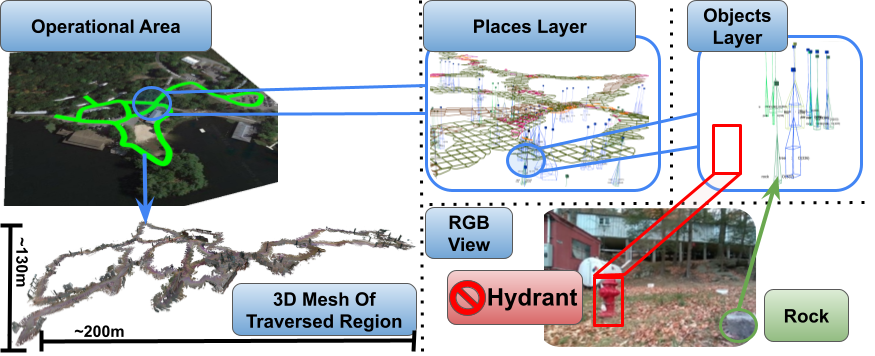}
    \caption{An outdoor 3D scene graph, where the operational area traversed by the robot is approximately 1$km$. An example generated mesh is shown on the left below the robot's trajectory, with zoomed-in selections of key parts of the graph's ontology. The graph has 314 objects, 15944 places, and 124 regions. Encoding this entire graph in the context window of an LLM exceeds the context window size of most contemporary models. Moreover, if we want the robot to follow a command to navigate to the fire hydrant, which is absent from the graph but near the known rock, we need a mechanism of using language to modify the scene graph to add the hydrant, which requires reasoning about where ``near the rock'' would be in the graph. That reasoning in turn involves identifying an object and a set of relevant places through a question expressed in natural language. Generating a plan requires reasoning about which parts of the scene graph are relevant for the navigation problem, which involves identifying the new object and a much larger set of relevant places. }
    \label{fig:scene-graph-image}
\end{figure*}

Consider the problem of directing a robot within the complex, outdoor environment shown in the overhead image on the left in Figure~\ref{fig:scene-graph-image}.
To navigate and act safely in such an environment, modern robot systems rely on structured world representations, such as voxel maps for collision avoidance, object detections for interaction, and spatial-semantic abstractions (e.g., regions) for efficient planning.
While effective, these perception-driven world models are abstractions of reality, are often noisy and incomplete, and capture only a subset of the information perceived by humans; this can lead to miscommunication.
Our goal is to enable users to give natural language commands to robots for tasks such as navigating to a goal or retrieving an object, which requires a shared representation between human understanding and the robot's model of the world.
We leverage natural language to bridge the gap between the robot's perception system and the human's understanding of that system.

There is a rich history of mapping from natural language to symbols grounded in bespoke structured environment models~\cite{tellex2020robots}, such as for planning ~\cite{tellex2011understanding,what-to-do-and-how-to-do-it,seq2seq-grounding} or embodied question answering~\cite{das2018embodied,azuma2022scanqa}; however these approaches are often specific to a particular task and are brittle with respect to the scope of language that can be handled.
More recently, 3D Scene Graphs (3DSGs)~\cite{armeni20193d,Hughes24ijrr-hydraFoundations,Gu24icra-conceptgraphs} and large language models (LLMs) \cite{achiam2023gpt} have become popular choices for representing the world and handling very general natural language, respectively.

3DSGs are hierarchical graph structures grounded in an ontology that defines concepts such as objects, places, and rooms.
Each node is typically assigned a symbol and stores relevant information about the entity, such as a position or a semantic class.
Hierarchical relationships are represented as edges.
Large-scale 3DSGs can contain thousands of nodes, such as the one in Figure~\ref {fig:scene-graph-image}.
To leverage them for task-driven navigation, an operator must understand the concepts captured in the different layers, such as the \emph{Objects} and \emph{Places} layers.
However, direct inspection of a graph is challenging for humans.
Therefore, our objective is to enable an operator to direct a robot towards a goal using natural language about the world rather than symbolic concepts of the graph. This raises the key question of how LLMs leverage 3DSGs to ground natural language instructions and queries.

Recent works that integrate an LLM with a 3D scene graph typically do so by populating the LLM's context window with a serialized text representation of the graph, such as a JSON string~\cite{strader2025language,Gu24icra-conceptgraphs,chang2025ashita}.
This technique works well for small, simple scene graphs, but, for several reasons, it does not scale to large or rich 3D scene graphs, like those that can be constructed by systems such as Hydra~\cite{Hughes24ijrr-hydraFoundations}.
Consider the kilometer-scale 3DSG shown in Figure~\ref{fig:scene-graph-image}, containing 314 objects, 15944 places, and 124 regions.
First, the full serialized text of this scene graph contains roughly seventeen million tokens, which exceeds the length of the context window of any contemporary models.
Second, even simple natural language expressions like ``the bag in the road" may require attending to distant parts of the context window in order to associate objects with the region they are contained in, which remains a challenge for even the most recent LLMs \cite{shi2023large,liu2023lost}.
Third, some scene graph ontologies are useful for robots but are idiosyncratic, making inter-layer associations in natural language more difficult to ground.
Finally, serializing the scene graph relies on the LLM to handle quantitative reasoning, such as answering the question ``which bag is closest to the robot?", for which LLMs are not well suited \cite{hendrycks2021measuring}.

Rather than include the full scene graph, several recent works proposed a pre-processing step during which a task-conditioned subgraph is generated~\cite{Rana23corl-sayplan,robot-navigation-based-on-3d-scene-graphs,booker2024embodiedrag}.
One approach uses an LLM and the hierarchical structure of the scene graph to enable broad filtering of irrelevant nodes~\cite{Rana23corl-sayplan}; however, these LLM-defined filters often result in multiple expensive interactions with the LLM's search for a sufficient subgraph.
Another approach takes inspiration from vector-based retrieval-augmented generation (RAG) \cite{gao2023retrieval}, encoding individual elements of the scene graph as vectors to be retrieved for context as needed, but it is susceptible to known vector-based RAG drawbacks, such as lacking sufficient context for multi-hop reasoning.
Neither approach directly addresses limitations related to the LLM's ability to ground language that requires quantitative reasoning~\cite{hendrycks2021measuring,cobbe2021training}.

We argue that a structured, graph-based query language can serve as a useful interface between an LLM and a 3D scene graph encoded in a graph database.
GraphRAG~\cite{han2024retrieval} is an extension of RAG techniques to graph data structures.
We propose using GraphRAG to interface LLMs with 3D scene graphs by (1) encoding a scene graph, like that produced by Hydra~\cite{Hughes24ijrr-hydraFoundations}, in a graph database and (2) providing an LLM with an interface to the database that accepts queries encoded in a graph query language called Cypher~\cite{cypher}.
Token efficiency is improved by selectively retrieving scene graph data as needed to ground a given user instruction or question, permitting scaling to large, rich graphs like that shown in Figure~\ref{fig:scene-graph-image}.
In addition, Cypher supports both graph-based relational queries and geometric spatial indexing, permitting the offloading of some or all of the quantitative aspects of language grounding that an LLM may struggle with.

Recent efforts use LLMs as agents that have access to tools, termed ``agentic AI", and show improved performance on reasoning tasks~\cite{patil2024gorilla,schick2023toolformer}.
We propose treating the Cypher-based query interface as a tool that the LLM can choose to use as a form of GraphRAG in support of grounding natural language instructions and questions to the 3D scene graph.
In this way, the LLM is able to correct query failures, try new queries in cases of insufficient information, or construct database response-dependent queries in a multi-query fashion, among others.

We evaluate our proposed approach through extensive experiments on both a small indoor and a large outdoor scene graph for the tasks of instruction grounding and scene question-answering.
We compare against an existing text-serialization approach as a baseline.
We also compare against a novel code generation baseline in order to assess if a generic API is sufficient for subgraph retrieval, or if the special-purpose database queries provide a more efficient mechanism for retrieval.
We show that our Cypher-based interface uses the fewest tokens and also has a higher success rate on both tasks.
To show the utility of our approach for robot systems, we also describe a physical demonstration of a realistic, complex object-retrieval scenario; this demonstration can be seen in a supplemental video. Code for the paper will be released after double-blind review.

\section{Background} \label{sec:background}
We provide preliminary information about 3D scene graphs, including the ontology we use, and Cypher, the graph database query language used by the LLM to retrieve scene graph data.

\subsection{3D Scene Graphs} \label{subsec:background-3dsg}
3D scene graphs provide a structured way to store metric, semantic, and topological information about the environment.
A 3D scene graph is a hierarchical graph $\mathcal{G}$ with vertices $\mathcal{V}$ and edges $\mathcal{E}$.
The vertices are partitioned into layers of increasing sparsity, with $\mathcal{V}_i$ denoting vertices in the $i^{th}$ layer.
Lower layers of the scene graph represent fine-grained spatial information such as meshes and individual objects, while higher layers of the scene graph represent more abstract concepts such as rooms for indoor environments (e.g., a kitchen) or regions for outdoor environments (e.g., a courtyard).
Each layer of the graph is composed of a set of nodes, each of which has a position and a list of layer-specific attributes, such as semantic class or bounding box.
Nodes are connected by intralayer edges that represent relative spatial relationships and interlayer edges that represent containment.

Building a scene graph requires specifying the ontology, that is, the construction of each layer of the graph and the set of possible attributes attached to each node and edge in the graph.
We denote a specific scene graph ontology as $\Upsilon$.
A recent survey presents an overview of many proposed 3D scene graph ontologies and methods for constructing them~\cite{catalano20253d}.
In this work, we adopt a subset of the ontology proposed in Hydra~\cite{Hughes24ijrr-hydraFoundations}, although our approach is flexible and does not depend on a specific set of layers, edges, or nodes and their properties.
Our ontology consists of four layers:
\begin{itemize}
    \item The \textit{Objects} layer stores objects as nodes with a semantic label, a position, and a bounding box. Edges between objects can capture relational properties, like \textit{on top of}.
    \item The \textit{Mesh Places} layer consists of connected, small (roughly 1m area) reachable, polygonal places stored as nodes that a robot can traverse through. Each node has a semantic label, a position, and a set of boundary points. Edges between places capture traversable connectivity for a robot. The Mesh Places layer is above the Objects layer, and interlayer edges represent the containment of objects in their respective, connected places.
    \item The \textit{3D Places} layer consists of connected 3D reachable places stored as nodes with a position, and a bounding sphere representing free space.
    \item The \textit{Room} or \textit{Region} layer stores geometric and semantic abstractions as nodes that represent higher-level spatial-semantic concepts. Each node has a position and a semantic label. Intralayer edges represent abstract traversable connectivity, indicating connectivity in the lower Mesh Places layer. Interlayer edges represent the containment of mesh places by a room or region.
\end{itemize}

\subsection{Cypher for 3D Scene Graph Queries} \label{subsec:background-cypher}
Cypher~\cite{cypher} is a declarative graph query language for specifying queries and update operations on graph-structured data, and these operations are usually executed by a graph database~\cite{graph_database_survey} storing the data of interest.
The Cypher data model considers a graph of nodes, each with a label (e.g., \texttt{Object}) and a set of key-value attribute pairs (e.g., \{class: vehicle\}).
Typed edges may be added between pairs of nodes to indicate relationships.
We can therefore instantiate a 3DSG in a graph database and interface with it via Cypher.

While there are many useful features of Cypher, we highlight examples that help illustrate how Cypher can be used to query 3D scene graphs and can help with certain steps an LLM might need to take when grounding natural language inputs from a user\footnote{A detailed introduction to Cypher can be found at \url{https://neo4j.com/docs/cypher-manual/current}.}.
First, a user may want to query the robot for information about the world, such as how many objects of a particular type are in certain parts of the world; for example, ``I was told there were some bags of supplies lying nearby. How many did you see?"
Cypher provides clauses that can return entries that \textit{match} a specific type and also a \textit{count} of the occurrence, making such a query simple to construct.

Second, since 3D scene graphs are primarily intended for use by a robot, the ontological hierarchy may not be well aligned to the expectations of an LLM (or a user).
If there are many bags in the world, our user might say ``The bags I am looking for should be in a courtyard. Bring them to me."
In our case, a courtyard is a type of region, and the user wants the bags that are contained in a courtyard region.
However, \textit{objects} are contained by \textit{places} which are contained by \textit{regions}, so determining the relationship between the bags and the courtyard requires traversing through the \textit{places} layer.
Cypher can decouple the query logic from this implicit multi-layer traversal via a simple syntax to find all transitive containment relationships.

Third, some kinds of natural language expressions use spatial relationships to unambiguously refer to specific objects and require quantitative reasoning to determine the correct referent; for example, ``Retrieve some more supplies from a nearby truck."
Cypher includes a concept of geometric \textit{points} and \textit{bounding boxes}, among others, and provides operations that can directly compute the distance between two points or whether a point is contained within a bounding box, for example.
These spatial operations could be used by an LLM to determine which bags are by a nearby truck, rather than having the LLM do this reasoning directly.

\subsection{Grounding Instructions to PDDL}
We want users to be able to specify instructions to a robot that it should then execute in the world.
Many contemporary instruction following methods translate the instruction to a formal planning representation and rely on an associated planner to find a corresponding plan.
While our approach is not specific to any planning representation, in
this work we use the Planning Domain Definition Language (PDDL)  \cite{Ghallab98-pddl}.
Specifically, we investigate the performance of an LLM on the task of translating from a natural language instruction to PDDL goal clauses grounded with symbols from a 3D scene graph.

PDDL allows for the specification of different parameterized planning domains with which to instantiate specific planning problems that fit a given domain description.
A PDDL problem instance is typically solved with a search-based solver such as Fast Downward~\cite{helmert2006fast}.
\textit{Symbols}\footnote{We use the term \emph{symbol} instead of the typical \emph{object} to avoid overloaded terminology with scene graphs.} are the entities in PDDL from which the parameters of the domain can be assigned a particular value.
These symbols are used to represent relevant features of a given domain, such as the rooms or objects in a scene graph for a mobile manipulation domain.
\textit{Predicates} in PDDL are Boolean functions that take a tuple of symbols as input.
Predicates are used to encode the state abstraction of the world, such as assigning types to symbols or relational properties to groups of symbols.
For example, the unary predicate \textsc{Holding(?obj)} can be used to indicate that an object is currently being held by a robot; or, the binary predicate \textsc{ObjectInPlace(?obj ?place)} can be used to indicate that an object is located in a particular place.
A \textit{goal} clause consists of grounded predicates defining a set of states that, if reached, constitute completing the task.
For example, consider a world with a bag object with the identifier \textsc{Bag1}.
The instruction ``Pick up the bag" could be encoded as the goal clause (\textsc{Holding Bag1}).
We therefore treat the problem of grounding natural language instructions as that of mapping to PDDL goal clauses consisting of predicates grounded with symbols from a 3D scene graph.

\section{Problem Statement} \label{sec:problem-statement}
Our goal is to enable LLMs to use 3D scene graphs for grounding natural language.
In particular, we are interested in grounding natural language that a user might provide as input to a robot system, and we focus on two specific tasks to make our goal concrete\footnote{We note that our approach is not specific to these tasks and is more broadly applicable to most tasks that require an LLM to use a 3D scene graph.}: (1) translating instructions to PDDL goals (i.e., to enable the robot to execute the instructions) and (2) answering questions about the scene (i.e., to inform the user of the robot's world knowledge).
We consider existing pre-trained LLMs and do not address methods that modify the model architectures or update the learned model weights.

For each language grounding task, we assume that the robot system has access to a 3D scene graph of its operating environment, denoted as $\mathcal{G}$.
For the task of translating instructions to PDDL, we assume that the robot system has access to a predefined PDDL domain $D$ that consists of predicates $P$ that are sufficient for encoding any of the provided instructions as a goal clause $G$.
A user provides a natural language instruction $l_{i}$ to the robot (e.g. ``Bring me the supplies from the courtyard."), and the LLM is responsible for generating a corresponding PDDL goal clause $G$ consisting of domain predicates that are grounded with symbols from the 3D scene graph:

\begin{equation} \label{eq:instruction-translation-task}
    LLM(l_{i};P,\mathcal{G}) \rightarrow G
\end{equation}

In the question answering task, a user provides a natural language question $l_{q}$ about the scene graph (e.g., ``how many supply bags are near the truck?"), and the LLM is responsible for interfacing with the scene graph data to generate the answer $a$:

\begin{equation} \label{eq:question-answering-task}
LLM(l_{q}; \mathcal{G}) \rightarrow a.
\end{equation}

In each task, the language provided by the user references the world, and the LLM must associate those references with the 3D scene graph in order to generate a response.
In this paper, we address the problem of how to provide the 3D scene graph data to the LLM in order to solve each of the above tasks.

\section{Technical Approach} \label{sec:technical-approach}
We assume access to a large language model $p_{llm}(X)$ that assigns probabilities to a string $X$ by decomposing it into a sequence of discrete \textit{tokens} $X=(x_1,\dots,x_N)$ and models the sequence's probability as a product of conditional next-token probabilities.
Given an input sequence $X$, an autoregressive LLM generates an output token sequence by sampling a token $x_{N+1}$ from the next-token distribution, appending it $X$, and repeating until a termination condition is reach (e.g., a special termination token is sampled as the next token).

In general, an LLM can be used to solve many different tasks encoded as text via \textit{prompting}, which involves providing an initial token sequence, or \textit{prompt}, in the LLM's context window that describes the task.
The quality of the generated response, therefore, depends on the provided prompt.
For tasks that depend on external knowledge (i.e., not part of the LLM's training set), retrieval-augmented generation (RAG) \cite{gao2023retrieval} has become a popular technique to dynamically retrieve the relevant information and add it into a prompt for the LLM to use as context.
This knowledge is retrieved as a token sequence $X_{rag}$ from some external knowledge source $K$ conditioned on some or all of the initial prompt $X_{prompt}$ via a \textit{retriever} function $r$:

\begin{equation} \label{eq:rag}
r(X_{prompt};K) \rightarrow X_{rag}
\end{equation}

For many RAG applications, $K$ is instantiated as a vector database of text embeddings $K = \{k_1,\dots,k_{|K|}\}$, and $r(\cdot)$ is a vector search process that embeds some or all of the prompt and retrieves the most similar embedding.
The retrieved token sequence is typically used to populate a templated part of the prompt.
In this work, we want to condition the LLM's output generation on the robot's world model, which is external knowledge represented as a 3D scene graph.

\subsection{Generating Cypher Queries} \label{subsec:technical-approach-cypher}

\begin{figure}
    \centering
    \includegraphics[width=1.0\linewidth]{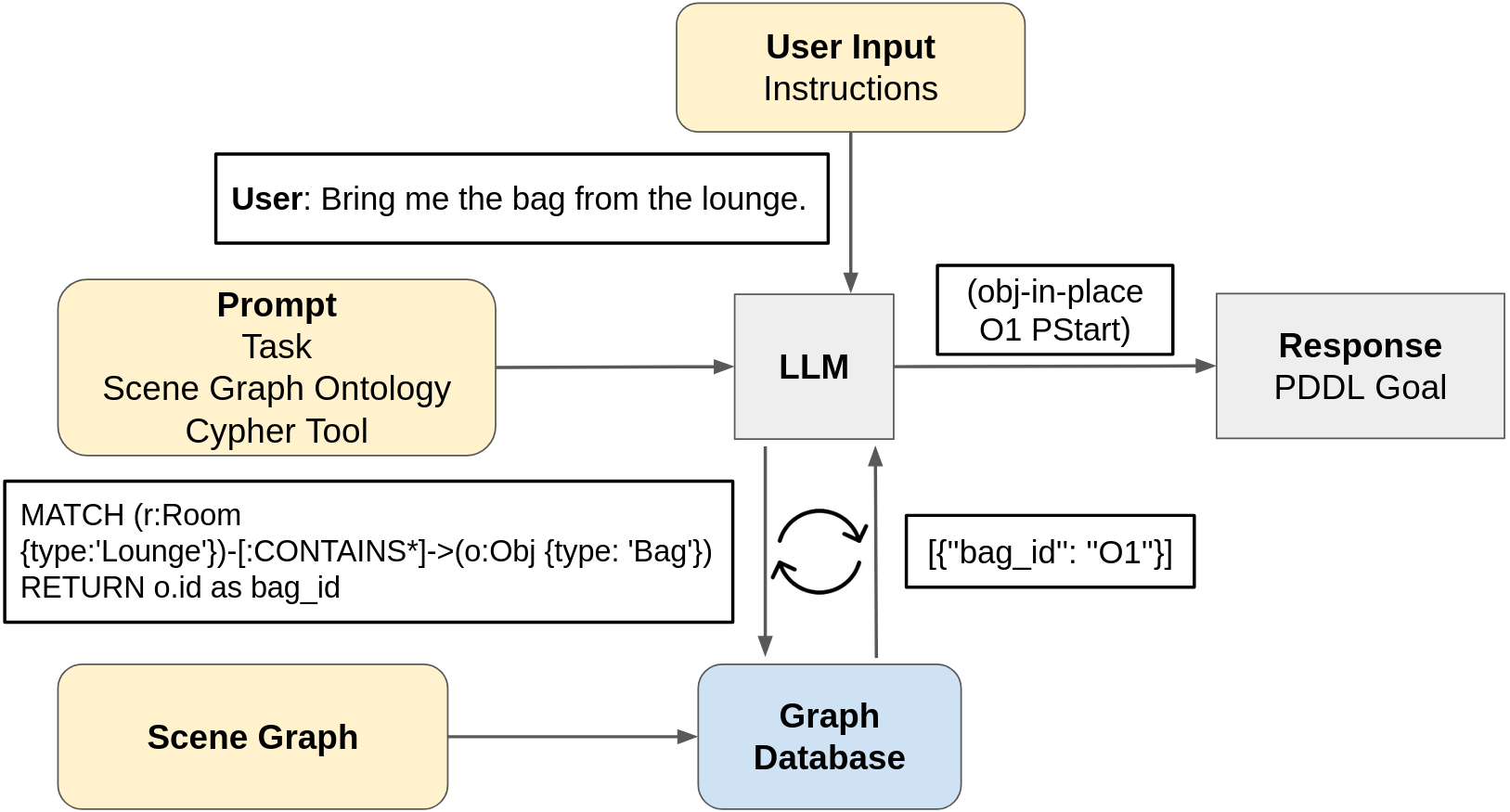}
    \caption{An illustration of the pipeline for using Cypher queries to a graph database as a tool for the LLM to use to retrieve relevant scene graph information. Given the prompt and the user input, the LLM chooses whether to interact with the graph database or generate a final response. The LLM can interact with the database zero or more times, up to some maximum number of interactions.}
    \label{fig:agentic-cypher}
\end{figure}

Rather than vector-based RAG, we take a GraphRAG \cite{han2024retrieval} approach by encoding the scene graph into a graph database that can be queried via Cypher.
For our retrieval mechanism, we model the Cypher-based interface as a tool for agentic LLM-driven tool use \cite{patil2024gorilla,schick2023toolformer,achiam2023gpt} and rely on the LLM itself to generate a Cypher query that, when executed on the database, returns the relevant external knowledge.

We can treat a single LLM-generated tool call as a pause in the autoregressive token generation process that is triggered by a special token.
Given the initial prompt $X_{prompt}$, the LLM will generate an initial output token sequence $X_o$ that consists of some initial reasoning chain $X_{o_{cot}}$, a Cypher query to execute $X_{o_{query}}$, and a special pause token $x_{pause}$: $X_o = (X_{o_{cot}},X_{o_{query}},x_{pause})$.
Once token generation is paused, we execute the Cypher query on the graph database for the LLM, which produces a response $X_{rag}$.
We then append the response $X_{rag}$ to the initial output token sequence $X_o$ and restart the token generation process, which is now conditioned on both the generated Cypher query and the database response.

We provide descriptions of the following in our initial prompt: the task $X_{task}$, the 3D scene graph onotology $X_{\Upsilon}$, an overview of Cypher as a tool $X_{cypher}$, and the input from the user $X_{user}$.
For translating an instruction to a PDDL goal clause, we also include a description of the available predicates $P$ in the PDDL domain $X_{P}$.
Notably, the LLM is responsible for deciding whether to generate a Cypher query or generate a final response to the user; therefore, the LLM may generate $[0,M)$ Cypher queries prior to providing a final response, where $M$ is a hyperparameter for the maximum number of tool calls.

Figure~\ref{fig:agentic-cypher} illustrates the prompt construction, agentic tool use for database interactions, and final response generation pipline.
Consider a robot operating in an indoor academic environment consisting of hallways and a lounge; there are several bags in the world but only one bag in the lounge.
The robot's scene graph is loaded into a graph database (bottom).
A user tells the robot ``Bring me the bag from the lounge" (top).
The instruction and the prompt materials are combined into the initial prompt ($X_{prompt}$) and given to the LLM.
The LLM generates a Cypher query to \textit{match} all instances of lounge type Rooms `r` that \textit{contain} a bag type Object `o` and \textit{return} the id of the object, if a match is found.
The database returns an id of ``O1" for the bag, and the LLM generates a PDDL goal (\textsc{obj-in-place O1 PStart}), where PStart is the 3D scene graph Place node of the current location of the user.
To reach this goal state, a PDDL planner will find a plan that involves the robot moving to the bag, picking it up, and bringing it back to the user.

\section{Experimental Design} \label{sec:experimental-design}
We evaluate the performance of several different approaches for interacting with and reasoning about scene graphs for instruction following and scene question-answering.
We construct evaluation datasets for each task using both a small indoor and large outdoor scene graph.

\subsection{Approaches} \label{subsec:experimental-design-approaches}

\textbf{Context Window}: As a baseline, we implement the standard approach of serializing the scene graph into structured text to be included in the LLM's context window.
Since the choice of which nodes and properties of the scene graph to include affects the kind and quality of response, we choose to include a subset that contains sufficient information with which to handle any natural language in our datasets.
We include the following:
\begin{itemize}
    \item \textit{Object Layer}: unique id, semantic label, position, parent 2D Places unique ids
    \item \textit{Place Layer}: unique id, sibling Places unique ids, parent Regions unique ids
    \item \textit{Region Layer}: unique id, semantic label, position
\end{itemize}

\textbf{Python Generation}: We implement a novel baseline that uses Python code generation to interface with the 3D scene graph.
In particular, we are interested in examining whether an existing API to scene graphs can be used for GraphRAG as we have constructed it, or if the special purpose retrieval operations provided by a graph database provides a useful inductive bias.
We provide the LLM with the available Python API for Hydra \cite{Hughes24ijrr-hydraFoundations} that consists of accessor functions and prompt it to generate user input-conditioned Python functions that retrieve and process the relevant scene graph data.
Like the Cypher querying, we model the code generation step as agentic tool use; the LLM is responsible for deciding whether to generate additional Python code or produce a final response to the user.
This returned content is included as context when generating the final response.

\textbf{Cypher Queries}: We implement our proposed approach, described in Section \ref{subsec:technical-approach-cypher}, to have the LLM generate Cypher queries to a Neo4j graph database containing the 3D scene graph and use the database response as context when generating the response to the user.

For both the Python code generation baseline and the proposed Cypher-based method, we also report results for an ablated version that enforces exactly one code generation or Cypher query step, respectively.

\subsection{Models} \label{subsec:experimental-design-models}
For each approach, we test the performance across several different proprietary frontier and open-weight, locally-hosted language models.
We keep the prompts consistent across models.
Below, we list the models used:
\begin{itemize}
   \item (OpenAI) \textsc{GPT-5}, \textsc{GPT-4.1}, \textsc{GPT-4.1-mini}, \textsc{GPT-4.1-nano}
   \item (Anthropic) \textsc{Claude Opus 4.1}
   \item (Alibaba) \textsc{Qwen3-32B}
\end{itemize}

We access the OpenAI models via the provided endpoints using the official Python API \cite{openai2025openai-python}.
Claude Opus 4.1 is a hybrid reasoning model.
We access it via Amazon Bedrock using the AWS Bedrock Python API~\cite{bedrock-api}, and we run it without \textit{extended thinking} mode.
Qwen3~\cite{Qwen3_2025} is an open-weight hybrid reasoning model.
We test Qwen3-32B with 4-bit quantization, running in \textit{thinking} mode.
The model runs locally on an Ollama~\cite{ollama2025ollama} instance running on a workstation with an NVDIA RTX4090 GPU.
We run all models with a 0 temperature setting.

\subsection{Scene Graphs} \label{subsec:experimental-design-scene-graphs}
We construct our evaluation datasets from two 3D scene graphs: a small indoor environment and a larger outdoor environment (Figure \ref{fig:scene-graph-image}).
Both scene graphs were built using the Hydra real-time scene graph library \cite{Hughes24ijrr-hydraFoundations} from RGBD sensor data (Intel Realsense D455 and ZED 2i cameras) collected via robot teleoperation on a Boston Dynamics Spot robot.
The small 3D scene graph is on the scale of 10s of meters and represents one floor of an indoor academic building.
There are 65 total objects over 12 different object categories, 96 total mesh places, and 5 total regions over 2 different region categories (hallway and lounge).
The large 3D scene graph is kilometer scale and represents an outdoor summer camp environment consisting of buildings, roads, tennis courts, lakefront, and rugged walkways.
There are 314 objects over 12 different object categories, 15,944 total 2D places, and 124 regions over 4 different region categories.

\subsection{Tasks} \label{subsec:experimental-design-tasks}
To investigate and compare the different approaches, we evaluate their performance when applied to the two language grounding tasks:

\begin{itemize}
    \item \textbf{Question-Answering}: Given a question in natural language about the scene graph, the LLM must generate an answer. An example question is ``Which cars are in parking lots?" The questions are designed such that their answers are either a Set, List, Dictionary (mapping from key to value), or Point.
    \item \textbf{Instruction Grounding (PDDL Translation)}: Given an instruction in natural language for a robot to execute, the LLM must translate the instruction to a PDDL goal grounded in the scene graph. An example instruction is ``Grab the bag and bring it to one of the parking areas." The instructions must be mapped to PDDL predicates that encode the goal. A goal is considered correct if it is logically equivalent to the ground truth solution. There is not necessarily a feasible plan to achieve the goal.
\end{itemize}

For each task, we construct an evaluation dataset that includes labeled examples associated with both the small and large scene graph.
The Instruction Grounding dataset consists of 82 instructions in the large 3DSG and 102 instructions in the small 3DSG.
The instructions vary whether the grounding references are directly referenced, use spatial relations for reference, or use the hierarchy of the scene graph for disambiguation.
The Question-Answer dataset consists of 100 queries of varying complexity for each scene graph.
We evaluate each approach on each dataset and report the success rate.

\section{Results \& Discussion} \label{sec:results-and-discussion}
\subsection{Task Success}
\begingroup

\begin{table}[]
\centering
\begin{tabular}{lcccc}
\toprule
& \multicolumn{2}{c}{Q\&A} & \multicolumn{2}{c}{PDDL} \\
\cmidrule(lr){2-3} \cmidrule(lr){4-5}
Method &  S & L & S & L  \\
\midrule
Cypher (A) (Ours)         & \textbf{0.85}    & \textbf{0.77}    & \textbf{0.65}    & \textbf{0.56}    \\
Cypher              & 0.82           & \textbf{0.77}           & 0.55           & 0.46           \\
Context Window      & 0.75        & 0.33        & \textbf{0.65}        & 0.41        \\
Python (A)          & 0.51    & 0.32    & 0.38    & 0.51    \\
Python              & 0.38           & 0.31           & 0.35           & 0.55           \\

\bottomrule
\end{tabular}
\caption{Task success rate for the different approaches. All results are for GPT-4.1. Methods with ``(A)" indicate the agentic version. ``S" and ``L" refer to the small and large scene graph, respectively.\vspace{-.45cm}}
\label{tab:success-rate-all-methods}
\end{table}

\endgroup Table~\ref{tab:success-rate-all-methods} reports the success rate of each approach on both the question-answering task (Q\&A) and the instruction grounding task (PDDL) for both the small (S) and large (L) scene graph.
Results are for a single model (GPT-4.1).

We note several interesting results.
First, our proposed agentic Cypher-based method either matches or exceeds the performance of all other methods for both tasks on both scene graphs.
Second, the context window baseline is a competitive method on small scene graphs, outperforming the Python-based approach on both tasks, and matching the performance of the agentic Cypher-based method on the PDDL task.
Our method significantly outperforms the context window baseline on the Q\&A task, however.
The context window baseline is also significantly worse than our method for both tasks on the large scene graph.

The agentic versions of both the Cypher-based interface and the Python code generation methods generally outperform their non-agentic ablations, indicating that treating the retrieval step of GraphRAG as a tool is a useful component of the method.
However, the degree of improvement due to tool use is varied.
For example, the code generation method's performance improved by 13\% on the Q\&A task for the small scene graph, but only improved by 1\% on the large scene graph (and even underperformed the non-agentic ablation for the PDDL task on the large scene graph).

Qualitatively, we find that all methods struggle with language that requires traversing the hierarchy of the scene graph.
While Cypher does provide a transitive containment operation, the GPT-4.1 inconsistently chooses to generate queries that take advantage of it.
We also notice that the Python method struggles to reliably use the correct API for certain functions, which is a consequence of using an existing API that is not optimized for LLM use.
In contrast, Cypher is a well-established language with many examples that are likely to have been included in the language model's pre-training data.

\begingroup

\begin{table}[h!]
\centering
\begin{tabular}{llcccc}
\toprule
& & \multicolumn{2}{c}{Q\&A} & \multicolumn{2}{c}{PDDL } \\
\cmidrule(lr){3-4} \cmidrule(lr){5-6}
Method & Model & S & L & S & L \\
\midrule

\multirow{3}{*}{{\rotatebox[origin=c]{0}{\makecell{Cypher (A) \\ (Ours) }}}}
  & GPT-4.1      & \textbf{0.85} & \textbf{0.77} & \textbf{0.65} & 0.56                 \\
  & GPT-4.1-mini & 0.69 & 0.65 & 0.60 & \textbf{0.61} \\
  & GPT-4.1-nano & 0.08 & 0.07 & 0.13 & 0.15 \\
  \midrule

\multirow{3}{*}{{\rotatebox[origin=c]{0}{\makecell{Context \\ Window}}}}
  & GPT-4.1      & 0.75     & 0.33     & \textbf{0.65}     & 0.41     \\
  & GPT-4.1-mini & 0.58 & - & 0.48 & - \\
  & GPT-4.1-nano & 0.03 & - & 0.12 & - \\
  \midrule

\multirow{3}{*}{{\rotatebox[origin=c]{0}{\makecell{Python \\ (A)}}}}
  & GPT-4.1      & 0.51 & 0.32 & 0.38 & 0.51                 \\
  & GPT-4.1-mini & 0.43 & 0.35 & 0.33 & 0.37 \\
  & GPT-4.1-nano & 0.15 & 0.14 & 0.09 & 0.15 \\
  \midrule

\bottomrule
\end{tabular}
\caption{Task success rate across methods for different model sizes. The Cyper and Python methods are agentic. ``S" and ``L" refer to the small and large 3D scene graphs, respectively.\vspace{-.4cm}}
\label{tab:success-rate-model-size}
\end{table}

\endgroup Table~\ref{tab:success-rate-model-size} reports the success rate of the approaches across different sizes of GPT models.
We see the expected trend that performance decreases with model size.
Using GPT-4.1-mini, our approach is both competitive with the context window approach and outperforms the Python approach when they are using GPT-4.1.
GPT-4.1-nano achieves very poor performance across all methods.

\begingroup

\begin{table}[b!]
\centering
\begin{tabular}{lcccc}
\toprule
 & \multicolumn{2}{c}{Q\&A} & \multicolumn{2}{c}{PDDL}  \\
\cmidrule(lr){2-3} \cmidrule(lr){4-5}
Model & S & L & S & L \\
\midrule
Claude Opus-4.1 & \textbf{0.94}             & \textbf{0.83}             & \textbf{0.88}             & \textbf{0.80} \\
   GPT-5         & 0.86                    & 0.81                    & 0.77                    & 0.70 \\
   GPT-4.1       & 0.85                 & 0.77                 & 0.65                 & 0.56 \\
   Qwen3-32B     & 0.74 & 0.67 & 0.56 & 0.55 \\

\bottomrule
\end{tabular}
\caption{Task success rate of the proposed agentic Cypher method's accuracy across OpenAI and Anthropic flagship models and one state of the art open source model run locally.}
\label{tab:success-rate-flagship-models}
\end{table}

\endgroup Table~\ref{tab:success-rate-flagship-models} reports the performance of our approach across different ``flagship" models that are representative of the many available models, both proprietary and open-source.
Notably, Claude Opus 4.1 is the top performer across both tasks and scene graphs.
We also note that our approach performs well with Qwen3-32B, a small local model.
It is competitive with the GPT-4.1 context window baseline on the small scene graph Q\&A task (74\% vs 75\%), and outperforms on the large scene graph Q\&A task (67\% vs 33\%) and PDDL task (55\% vs 41\%).

\begingroup

\begin{table}[h!]
\centering
\begin{tabular}{lcccccc}
\toprule
& \multicolumn{3}{c}{Small 3DSG} & \multicolumn{3}{c}{Large 3DSG} \\
\cmidrule(lr){2-4} \cmidrule(lr){5-7}
Method &  I & T & O & I & T & O  \\
\midrule
Cypher (A) (Ours)   & \textbf{2366}    & 205 & 319    & \textbf{2395}    & 1596 &  942    \\
Context Window      & 8735        & - & \textbf{134}        & 582202 & -       & \textbf{56}        \\
Python (A)          & 6161    & 82 & 390    & 6237 & 33 & 337    \\

\bottomrule
\end{tabular}
\caption{Average Input (I) and Output (O) token counts for each method, for the small and large 3D scene graphs. The Tool (T) columns counts the number of additional input tokens that are passed to the LLM as the output of tool calls.}
\label{tab:token-counts}
\end{table}

\endgroup
In Table~\ref{tab:token-counts}, we show the average number of tokens used by each approach on the PDDL task.
We report the input tokens, which come from the prompt and the response of tool calls, and output tokens, which include tool calls and the final answer.
The serialized text encoding for the large 3DSG scales to an intolerable size for all models except the GPT-4.1 family.
Our method and the Python baseline have input token counts that are two orders of magnitude smaller and do not directly depend on the size of the 3DSG.
Our method uses the fewest input tokens, indicating efficient tool use and concise database responses.
While it is possible that a query \emph{could} result in exhaustive information about the 3DSG, our results show that, on average, the amount of data returned by the query execution is small compared to the full size of the scene graph; this is also true for the Python approach.
The Python baseline requires more input tokens compared to the proposed Cypher method, because it must represent the special-purpose scene graph API.
On the other hand, Cypher is a more general-purpose interface that is likely well represented on the internet and in LLM training data and requires less information within the prompt.

\subsection{Physical Robot Demonstration} \label{subsec:experimental-design-demo}
\begin{figure*}[t]
    \centering
    \includegraphics[width=\linewidth]{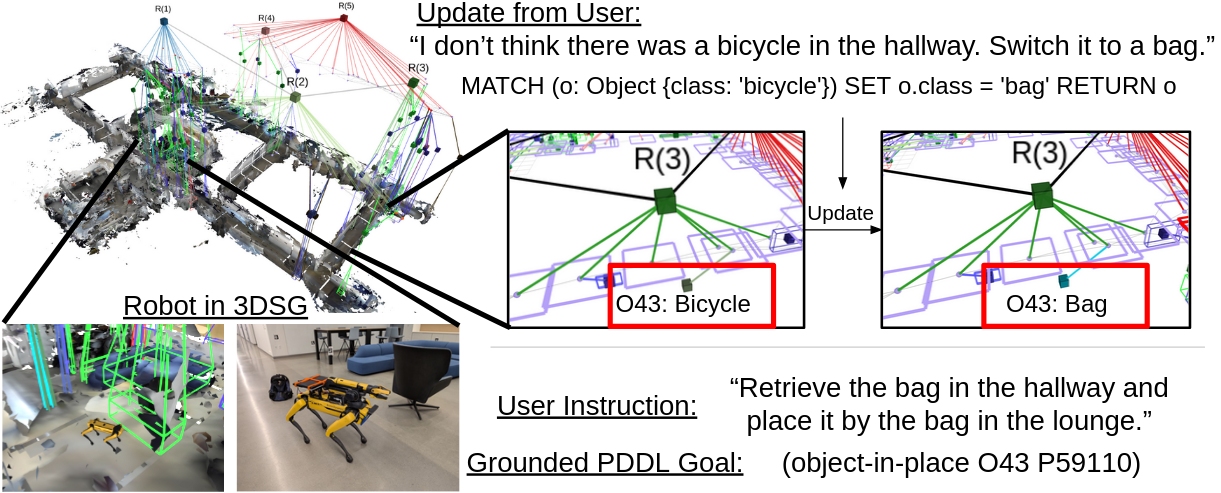}
    \caption{We demonstrate grounding a user's natural language instructions to a Cypher query to correct the object label in a 3DSG, producing a PDDL goal based on the updated object semantics, and then executing a plan to achieve the goal onboard a robot.}
    \label{fig:robot}
\end{figure*}
We also demonstrate our proposed Cypher-based scene graph interface on a Boston Dynamics Spot robot in the same indoor building that is represented by the small scene graph\footnote{\url{https://www.youtube.com/watch?v=zY_YI9giZSA}}.
The scenario, illustrated in Figure~\ref{fig:robot}, incorporates both questions and instructions to the robot, and also includes the user providing a natural language correction to the scene graph that is mapped to a corresponding Cypher query for database execution.
A user first asks the system several questions about the scene graph, such as the type, count, and location of objects, in order to improve their own understanding of what is available to the robot.
Upon noticing that a bag in the scene graph was mislabeled as a bicycle, the user tells the robot about the object's correct semantic class, and the robot updates its scene graph to incorporate the new information.
Finally, the user gives a command in natural language to retrieve the newly labeled bag, resulting in a translated PDDL goal that is solved by Fast Downward \cite{helmert2006fast}.
Given the resulting plan grounded in the scene graph, the robot executes the plan, retrieving the object for the user.
A video of this demonstration is included as supplemental material.

\section{Conclusion} \label{sec:conclusion}
In this work, we investigate how to enable LLMs to ground natural language instructions and questions to a 3D scene graph representing a robot's model of the world.
We propose a GraphRAG approach in which an LLM can use a Cypher-based interface to query a graph database for relevant 3DSG data.
This data is used as context by the LLM to generate either (1) a PDDL goal for an instruction or (2) an answer to a question about the scene.
We treat the Cypher interface as a tool and allow the LLM to choose how frequently to interact with the database before giving a final response to the user.
Through experiments using examples annotated for a small indoor and a large outdoor 3DSG, we show that our proposed approach outperforms both a Python generation and context window-based baseline on both tasks for both scene graphs, scaling significantly better to the large scene graph.
Our method is more amenable for use with small models.
Using Qwen3-32B, it is competitive with the context window baseline using GPT-4.1 for the tasks on the small scene graph, and it significantly outperforms on the large scene graph.
By being more token efficient, our proposed Cypher-based agentic approach enables LLMs to ground natural language instructions and queries to rich, kilometer scale outdoor scene graphs.

One limitation of our proposed approach is that the underlying graph ontology must be included as part of the prompt to the LLM in order for it to generate grounded Cypher queries.
While this is not an issue for a static ontology, it may be useful for the concepts in the graph to change during robot operation.
Therefore, we are interested in techniques that permit updating and maintaining the description of the scene graph ontology to enable seamless use within or across sessions of operation; more generally, we are interested in maintaining good prompt descriptions of tools whose interfaces or function signatures may either change over time or are initially poorly described.

\section*{Acknowledgments}

This work was sponsored by the Army Research Laboratory under Cooperative Agreement
Number W911NF-17-2-0181.
L. Carlone holds concurrent appointments at MIT and as an Amazon Scholar. This paper describes work performed at MIT and is not associated with Amazon.
We thank Nina Polshakova for her instrumental work on the initial software implementation connecting 3D scene graphs to a graph database and generation of the initial set of questions and answers for developing this system.

\bibliographystyle{ieeetran}

\appendix
\subsection{Code Release}

We release the core infrastructure code for interfacing between 3D scene graphs and implementing the agentic tool-calling pipelines as free and open source code under the GPL-3.0 license. The connection between the \texttt{spark\_dsg} library and Neo4j graph database can be found at \url{https://github.com/GoldenZephyr/heracles}. The agentic workflow that supports tool calling and evaluation can be found at \url{https://github.com/GoldenZephyr/heracles_agents}. The planning and execution framework demonstrated in the paper can be found at \url{https://github.com/MIT-SPARK/Awesome-DCIST-T4}.

\subsection{Solution Comparison Methodology}

To evaluate if an answer is correct, we need to define a sense of equality. This
is rather tricky, because there are different senses in which things can be
equal. We implement separate equality checks for the question answering task and the
PDDL grounding task.

\subsubsection{PDDL}
PDDL goal strings define a set of world states that represent satisfaction of the goal. Two PDDL goals are equivalent if they describe the same set of states. Since individual PDDL facts can be composed with arbitrary nesting of conjunction, disjunction, and negation, and the goal set is invariant to reordering of facts combined in a conjunction and disjunction, two PDDL goals cannot be directly compared using string equality. Two PDDL goals should be considered equivalent if they are logically equivalent. In general, this requires a brute force comparison of truth tables. However, for the more limited set of goals considered in our evaluation questions, it is sufficient to verify that the disjunctive normal forms of each expression contain the same terms.

\subsubsection{SLDP}
The questions included in our question answering dataset are formulated to be naturally answered with a primitive (e.g., string, number, point) or a collection (e.g., list, set, dictionary). We define a concise syntax for the LLM to format its answers, with a well-defined semantics for comparing equality. We call this syntax and semantics for equality checking the SLDP (Set, Line, Dictionary, Point) language.

Numbers and strings are primitive expressions in the SLDP language. They are represented without any surrounding quotation. Points are also primitive expressions, represented as \texttt{Point(x y z)}. Sets are represented by a sequence of expressions within an enclosing pair of angle brackets. Lists are represented by a sequence of expressions within enclosing square brackets. A dictionary is represented by colon-separated key-value pairs, surrounded by curly braces. Listing 1 shows an EBNF grammar for the SLDP language, which we implement in Lark~\cite{lark-repo}.

\begin{listing}
\begin{minted}{ebnf}
expression = num | string | set | list
             | dict | point ;
string = CNAME ;
num = SIGNED_FLOAT | INT ;
set = "<" [ expression
            { "," expression } ] ">" ;
list = "[" [ expression
            { "," expression } ] "]" ;
kv = string ":" expression ;
dict = "{" [ kv { "," kv } ] "}" ;
point = "POINT(" num num num ")" ;
\end{minted}
\caption{EBNF definition for the SLDP equality comparison language. The CNAME terminal matches alphanumeric strings with a leading letter (valid identifiers in C), and SIGNED\_FLOAT and INT match floating point nubmers and integers respectively. Whitespace is ignored when parsing the language.
}
\end{listing}

Two strings are equal if they match exactly. Two numbers are equal if the absolute value of their difference is within $\epsilon$ (here 0.01). Two points are equal if their distance (in some norm) is at most $\delta$ (we use the $l_{\infty}$ norm and $\delta$ = 0.01).

A list is an ordered sequence of expressions. Two lists are equal if each pair of corresponding elements is equal.  A set is an unordered group of expressions. Two sets A and B are equal if A $\subseteq$ B and B $\subseteq$ A. A dictionary is a mapping of strings to expressions. Dictionaries
are equal if the sets of their keys are equal and the value for each key
matches between dictionaries, and two points are equal if they are  within some
tolerance. We support arbitrary compositions of these containers, although in practice none of our evaluation questions require deeply nested answers.

\subsection{Dataset Examples}
There are four distinct datasets used in the experiments for this work: one for each scene graph \& language task pair (e.g., the large scene graph paired with the PDDL translation task).
Within each dataset, we aimed to include a variety of language complexity that reference or rely on different concepts, such as containment in scene graphs (e.g., ``the light in the hallway") or conjunctions in PDDL (e.g., ``Inspect each garbage can").
Each question or instruction is chosen to have an unambiguous correct answer.
Below, we include 5 representative examples from each dataset (listed as 1-20) to help illustrate the kind of questions and instructions we evaluated.

\subsection*{Large Scene Graph \& Question-Answering}
\noindent \textbf{Example 1}\\
\textbf{Question}: What kind of objects are in the scene graph? \\
\textbf{Answer}: \textsc{\textlangle tree, fence, vehicle, seating, window, sign, pole, door, box, trash, rock, bag\textrangle}

\medskip
\noindent \textbf{Example 2}\\
\textbf{Question}: Which two rooms are farthest apart?\\
\textbf{Answer}:\textsc{\textlangle R30, R83\textrangle}

\medskip
\noindent \textbf{Example 3}\\
\textbf{Question}: There is a rock on the road. Which rock is it? \\
\textbf{Answer}: \textsc{O128}

\medskip
\noindent \textbf{Example 4}\\
\textbf{Question}: Which poles are within 5 meters of a fence? \\
\textbf{Answer}: \textsc{\textlangle O95, O99, O102, O381\textrangle}

\medskip
\noindent \textbf{Example 5}\\
\textbf{Question}: What is the average distance between an object and its closest objects that isn't the same type? \\
\textbf{Answer}: \textsc{60.00}

\subsection*{Small Scene Graph \& Question-Answering}
\noindent \textbf{Example 6}\\
\textbf{Question}: How many of each object type are there? \\
    \textbf{Answer}: \textsc{\textbraceleft seating: 22, sign: 8, storage: 15, food: 1, appliance: 2, decor: 5, trash: 4, bicycle: 1, box: 3, light: 2, bed: 1, bag: 1\textbraceright}

\medskip
\noindent \textbf{Example 7}\\
\textbf{Question}: Which room has the most neighbors?\\
\textbf{Answer}: \textsc{R2}

\medskip
\noindent \textbf{Example 8}\\
\textbf{Question}: Which box is closest to the bicycle? \\
\textbf{Answer}: \textsc{O59}

\medskip
\noindent \textbf{Example 9}\\
\textbf{Question}: Where are all of the trash can positions? \\
\textbf{Answer}: \textsc{\textlangle POINT(-18.70 -4.21 0.12), POINT(-19.22 -4.42 0.03), POINT(-20.96 -20.89 -0.02), POINT(-25.17 -22.58 -0.21)\textrangle}

\medskip
\noindent \textbf{Example 10}\\
\textbf{Question}: How many objects are in mesh places that are within 6 hops of the bag's mesh place? (Not counting the bag itself.) \\
\textbf{Answer}: \textsc{28}

\subsection*{Large Scene Graph \& PDDL Translation}
\noindent \textbf{Example 11}\\
\textbf{Question}: I want you to get a good look at the trash.\\
\textbf{Answer}: \textsc{(safe O54)}

\medskip
\noindent \textbf{Example 12}\\
\textbf{Question}: There is one trash can. Inspect it and then go to the window. \\
\textbf{Answer}: \textsc{(and (safe O54) (at-object O21))}

\medskip
\noindent \textbf{Example 13}\\
\textbf{Question}: Pick up the box that is closest to the bag. \\
\textbf{Answer}: \textsc{(holding O39)}

\medskip
\noindent \textbf{Example 14}\\
\textbf{Question}: There is a box in the road. Reposition the bag to be in the same place as that box. \\
\textbf{Answer}: \textsc{(object-in-place O85 P12659)}

\medskip
\noindent \textbf{Example 15}\\
\textbf{Question}: Bring one of the boxes to the same place as the farthest seating. \\
\textbf{Answer}: \textsc{(or (object-in-place O39 P3522) (object-in-place O55 P3522) (object-in-place O395 P3522) (object-in-place O397 P3522))}

\subsection*{Small Scene Graph \& PDDL Translation}
\noindent \textbf{Example 16}\\
\textbf{Question}: Go to object 79.\\
\textbf{Answer}: \textsc{(visited-object O79)}

\medskip
\noindent \textbf{Example 17}\\
\textbf{Question}: Ensure the bag is harmless. \\
\textbf{Answer}:  \textsc{(safe O285)}

\medskip
\noindent \textbf{Example 18}\\
\textbf{Question}: Some mesh places aren't in a room. Visit each of them. \\
\textbf{Answer}: \textsc{(and (visited-place P2441) (visited-place P3107) (visited-place P15561) (visited-place P25023) (visited-place P25697))}

\medskip
\noindent \textbf{Example 19}\\
\textbf{Question}: Grab the bicycle, or visit the hallways without going into the lounge. \\
\textbf{Answer}: \textsc{(or (holding O43) (and (visited-room R2) (visited-room R3) (visited-room R4) (visited-room R5) (not (visited-room R1))))}

\medskip
\noindent \textbf{Example 20}\\
\textbf{Question}: Go check the light in the hallway and make sure it's safe. Or bring all of the trashes to mesh place 1833. \\
\textbf{Answer}: \textsc{(or (safe O300) (object-in-place O19 P1833) (object-in-place O30 P1833) (object-in-place O64 P1833) (object-in-place O79 P1833))}

\subsection{Agentic vs. Non-Agentic Pipelines}
Our proposed method is to provide a scene graph interface (e.g., Cypher or a Python API) as a tool for an LLM to use and permit the LLM to decide when and how often to use the tool (up to a maximum number of tool calls).
By allowing multiple tool calls, it is possible to avoid initial failures via correction for issues like syntactically invalid Cypher or Cypher that incorrectly refers database labels.
It also allows for chaining tool calls in order to collect more information, if needed.
To evaluate the impact of this ``agentic ai" tool use, we implemented a variant of the method that includes exactly one tool call (See Section~\ref{sec:results-and-discussion} for the results).
The most significant difference in the implementation of these ``non-agentic" ablations is that we decompose the prompting of the LLM into two distinct steps:
\begin{enumerate}
    \item Prompt the model to generate a single tool call
    \item Prompt the model to generate a final response to the user given the response of the tool call
\end{enumerate}

\clearpage \onecolumn \subsection{Prompt Description}

Below we provide an overview of the prompts used in our experimentation.
Each individual prompt is composed of distinct components.
We describe these components and provide examples of their content.
We also fully reproduce a prompt for a single variant of the pipeline (Agentic Cypher for translating instructions to PDDL) to illustrate how these components can be composed into a single prompt for a given pipeline and task.

\subsection*{Prompt Skeleton}

While the prompt content for each combination of scene graph interface (e.g., Cypher) and task (question answering vs. PDDL goal grounding) is different, there are many parts of the prompt that can be shared. The following skeleton outlines the components that are composed into each prompt:\\
\textbf{System}: You are a helpful assistant who is an expert at...\\
\textbf{Scene Graph Description}: An overview of the Hydra 3D scene graph ontology.\\
\textbf{Labelspace Description}: A list of the semantic labels in the scene graph.\\
\textbf{Interface Description}: An overview of the scene graph interface (Cypher vs. Python vs. In-context)\\
\textbf{PDDL Domain Description}: Introducing the PDDL domain that should be grounded to.\\
\textbf{In Context Examples}: A list of example user inputs and LLM responses.\\
\textbf{User Input}: The new user input from a specific question\\
\textbf{Answer Formatting Guidance}: Explains how the LLM should format the output (e.g., as a PDDL goal between two \textlangle answer\textrangle tags)

Each variation of the prompt uses the same description of Hydra's scene graph ontology. The large and small scene graph each have their own labelspace. The interface description depends on whether Python, Cypher, or an in-context encoding is providing scene graph information to the LLM. The PDDL domain description is only included for the PDDL grounding task. The in-context examples depend on the task. The answer formatting guidance for question and answering presents an overview of the SLDP comparison language and tells the LLM what type of SLDP expression it should return.

\subsection*{Full Prompt Example}

We produce the full prompt for the agentic PDDL task, using Cypher as the scene graph interface. This prompt is identical to the prompt used in the agentic Cypher-based PDDL grounding experiments on the large 3D scene graph, except for formatting of newlines to fit in this form factor.

\begin{alltt}

\textbf{developer}: You are a helpful assistant who is an expert at assigning
robots PDDL planning goals based on natural language commands grounded in 3D
scene graphs. You have access to a database representing a 3D scene graph, which
stores spatial information that a robot can use to understand the world. Given a
command, your task is to generate a Cypher query that queries the relevant
information from the database. Then, use this information to formulate a PDDL
goal for a robot. Given a 3D Scene Graph and an instruction, use the provided
tool to query information from the scene graph. When you have enough
information, submit your final answer. You can call the cypher query tool up to
5 times. Explain your reasoning before each Cypher query. When producing a final
answer, follow the format and be concise.

\textbf{developer}: <Scene Graph Description> A 3D scene graph is a hierarchical
graph consisting of layers that each contain nodes. Graphs have the 2D Places
layer, the 3D Places layer, the Objects layer, and the Rooms layer.

3D Places Layer (PLACES): The 3D Places layer in the scene graph contains places
that are reachable locations in the world. Each Place node has a unique ID
(p<id>). Note that the unique ID uses a lowercase 'p' for the 3D Places layer.
The term for a node in this layer is a 'Place' node, NOT a 'Mesh Place' node.

2D Places Layer (MESH_PLACES): The 2D Places layer contains Places that are
reachable locations in the world with a possible semantic class (type). Each
Place node has a unique ID (P<id>). Note that the unique ID uses an uppercase
'P' for the 2D Places layer. The term for a node in this layer is a 'Mesh Place'
node, NOT a 'Place' node.

Objects Layer (OBJECTS): The Objects layer contains Objects that exist in the
world. Each Object has a unique ID (O<id>), a semantic class (type), an x,y,z
position (pos), and a parent. The parent can be a Place or a Mesh Place,
indicating which Place or Mesh Place the Object belongs to. Each Object will be
represented in the form: (id, type, pos, parent_place).

Rooms Layer (ROOMS): The Rooms Layer contains Rooms that exist in the world.
Each Room has a unique ID (R<id>), a semantic class (type), and an x,y,z
position (pos). Each Room will be represented in the form: (R<id>, type, pos).

A graph can have either a 2D Places Layer, or a 3D Places Layer, or both. The
hierarchy is strict: an Object's parent is always a Place or Mesh Place. A
Place's parent is always a Room. Therefore, to find all objects in a room, you
must first find all places that are children of that room, and then find all
objects that have one of those places as a parent. </Scene Graph Description>

\textbf{developer}: <Labelspace Description>
These are the labels available in the scene graph at each layer. Synonyms should
be mapped to a label in the the list.

<object_labels> tree vehicle signal rock fence boat sign door pole rail window
  flower bed box storage barrel bag basket seating flag decor light appliance
  trash bicycle food clothes </object_labels>

<room_labels> road field shelter indoor stairs sidewalk path boundary shore
  ground dock parking footing </room_labels>

<mesh_places_labels> water ground grass sand sidewalk dock path hill bridge
  wall floor stairs structure surface flora </mesh_places_labels>

</Labelspace Description>

\textbf{developer}: Labels in Database:
    - Object: a node representing an object in the world.
        - nodeSymbol: a unique string identifier
        - class: a string identifying the object's semantic class or type
        - center: the 3D position of the object, as a POINT type
    - MeshPlace: a node representing a 2D segment of space the robot might be
                 able to move to.
        - nodeSymbol: a unique string identifier
        - class: a string identifying the place's semantic class or type
        - center: the 3D position of the mesh place, as a POINT type
    - Place: a node representing a 3D region of free space
        - nodeSymbol: a unique string identifier
        - center: the 3D position of the place, as a POINT type
    - Room: a node representing a room or higher-level region
        - nodeSymbol: a unique string identifier
        - class: a string identifying the room's semantic class or type
        - center: the 3D position of the room, as a POINT type

Object, MeshPlace, Place, and Room are all Cypher labels attached to nodes.

Places and Mesh Places represent a higher level of the hierarchy compared to
objects, but lower level than rooms.

There are two kinds of existing edges. First is (a)-[:CONTAINS]->(b), which
connects nodes between different layers and means that b is contained within a.
Nodes in higher levels of the hierarchy may contain nodes in lower levels of the
hierarchy, but nodes in the lower level of the hierarchy will not contain
higher-level nodes. The other kind of edges represent connectivity within a
layer: [:OBJECT_CONNECTED], [:PLACE_CONNECTED], [:MESH_PLACE_CONNECTED],
[:ROOM_CONNECTED]. Remember that (a)-[:CONTAINS*]->(b) will match transitive
relationships.

Note that in the current version of cypher, `distance` has been replaced by
`point.distance`. Also, do not use any apoc functions in your queries.

\textbf{developer}: <PDDL Domain>

    The PDDL domain consists of the following predicates described below. These
    predicates get parameterized by symbols from a 3D scene graph.
    (visited-place ?P): This predicate indicates that a robot must visit Place
    '?P' at some point, where '?P' is a placeholder for a Place ID. (at-place
    ?P): This predicate indicates that a robot must be at Place '?P', where '?P'
    is a placeholder for a Place ID. (visited-object ?O): This predicate
    indicates that a robot must visit Object '?O' at some point, where '?O' is a
    placeholder for an Object ID. (at-object ?O): This predicate indicates that
    a robot must be at Object '?O', where '?O' is a placeholder for an Object
    ID. (safe ?O): This predicate indicates that a robot must inspect Object
    '?O', where '?O' is a placeholder for an Object ID. (visited-room ?R): This
    predicate indicates that a robot must visit Room '?R' at some point, where
    '?R' is a placeholder for a Room ID. (in-room ?R): This predicate indicates
    that a robot must be at Room '?R', where '?R' is a placeholder for a Room
    ID. The 'at' and 'visited' predicates are useful for specifying locations
    for robots to go. When an instruction indicates an order, you should use
    (at-place ?P), (at-object ?O), or (in-room ?R) to specify the final goal and
    (visited-place ?P), (visited-object ?O), or (visited-room ?R) to specify the
    intermediate goals.

    (holding ?O): This predicate indicates the a robot must be holding an Object
    '?O', where '?O' is a placeholder for an Object ID. The 'holding' predicate
    is useful for specifying that a robot should pick up an object.

    (object-in-place ?O ?P): This predicate indicates that an Object '?O' must
    be located inside a Place '?P', where '?O' is a placeholder for an Object ID
    and '?P' is a placeholder for a Place ID. The 'object-in-place' predicate is
    useful for specifying that a robot should place an object somewhere. If the
    robot should move an object from one place to another, you should use the
    'object-in-place' predicate for the goal instead of the 'holding' predicate.

    You can compose PDDL goal predicates into more complex goals using the
    following operators: not: the 'not' operator negates the truth value of the
    predicate. For example '(not (visited-room R1))' means Room 'R1' should not
    be visited. and: the 'and' operator is used to indicate that multiple
    predicates must be true. For example '(and (visited-room R1) (visited-object
    O1))' means both Room 'R1' and Object 'O1' must be visited. or: the 'or'
    operator is used to indicate that either one predicate or another predicate
    must be true. For example '(or (visited-room R1) (visited-object O1))' means
    that either Room 'R1' or Object 'O1' must be visited. If an instruction is
    not specific about which object, place, mesh place, or room that it is
    referring to, you MUST use the 'or' operator to construct a PDDL goal. This
    lets the planner determine which goal to achieve.

</PDDL Domain>

\textbf{developer}: The following is a set of examples of mapping instructions
to PDDL goals (assuming the results of a Cypher query).

\textbf{user}: Instruction: Head to place 100.
\textbf{assistant}: <answer>(visited-place P100)</answer>

\textbf{user}: Instruction: Move over to the trash.
\textbf{assistant}: <answer>(visited-object O15)</answer>

\textbf{user}: Instruction: Go to room 1.
\textbf{assistant}: <answer>(visited-room R1)</answer>

\textbf{user}: Instruction: Inspect the trash.
\textbf{assistant}: <answer>(safe O15)</answer>

\textbf{user}: Instruction: Pick up the box
\textbf{assistant}: <answer>(holding O120)</answer>

\textbf{user}: Instruction: Move the box to mesh place 11.
\textbf{assistant}: <answer>(object-in-place O120 P11)</answer>

\textbf{user}: Instruction: First pick up the box and then head over to the trash.
\textbf{assistant}: <answer>(and (holding O120) (at-object O15))</answer>

\textbf{user}: Instruction: Move to one of the poles on the sidewalk.
\textbf{assistant}: <answer>(or (visited-object O155) (visited-object O160)
(visited-object O185))</answer>

\textbf{user}: Instruction: Go pick up one of the poles on the sidewalk.
\textbf{assistant}: <answer>(or (holding O155) (holding O160) (holding O185))</answer>

\textbf{developer}: Now, generate a PDDL goal for this command:
\textbf{user}: Go to Object 1
\textbf{developer}: Make your answer as concise as possible
\textbf{developer}: Return the PDDL goal between two answer tags, e.g. <answer>
pddl goes here </answer>

\end{alltt}

\subsection*{In Context Scene Graph Interface Description}
\begin{alltt}
  <scene_graph_notes>
  Objects Layer: The Objects layer contains Objects that exist in the world.
  Each Object has a unique ID (id), a semantic class (type), an x,y,z position
  (pos), and a set of parent Places (parent_places). The parent Places indicate
  which Places the Object belongs to. Each Object will be represented in the
  form: (id, type, pos, parent_places).
  Places / Mesh Places Layer: The Places / Mesh Places layer contains Places
  that are reachable locations in the world. Each Place node has a unique ID
  (id), a set of sibling Places (siblings) and a set of parent Rooms
  (parent_rooms). Each Place will be represented in the form (id, siblings,
  parent_rooms).
  Rooms Layer: The Rooms Layer contains Rooms that exist in the world. Each Room
  has a unique ID (id), a semantic class (type), an x,y,z position (pos), and a
  set of sibling Rooms (siblings). Each Room will be represented in the form:
  (id, type, pos, siblings).
  </scene_graph_notes>
\end{alltt}

\subsection*{Python Scene Graph Interface Description}
\begin{alltt}
  <python_api>
  Your output should be executable python code (def solve_task(G) ) that takes
  in a 3D scene graph G. The result should be the information needed to answer
  the user's query. All nodes in the graph have a CASE SENSITIVE node symbol.
  For example, object 1 should be referenced as O1. Assume all python imports
  are present. Only give the python function and any helper functions you need.
  They will be automatically run. Ensure NodeSymbols are returned rather than
  bitmaps/uint64 when referencing symbols. Make sure to check the full
  hierarchical structure when looking at children and parent relationships.

  Here are examples of how to use the API to answer questions:
  1. Task: Get a list of all objects in the scene graph G
  Solution:
  <python>
  def solve_task(G):
      # Returns a list of all objects as dictionaries in the scene graph G
      objects = [o for o in G.get_layer(spark_dsg.DsgLayers.OBJECTS).nodes]
      return objects
  </python>
  2. Find all the foo in the graph G
  <python>
  def solve_task(G):
      # Returns a list of all foo (as dictionaries) in the scene graph G
      labelspace = G.metadata.get()["labelspace"]
      return [
          labelspace[str(o.attributes.semantic_label)]
          for o in G.get_layer(spark_dsg.DsgLayers.OBJECTS).nodes
          if "foo" in labelspace[str(o.attributes.semantic_label)].lower()
      ]
  </python>

  Please consider efficiency when writing your code. Use list comprehensions and
  avoid unnecessary loops. Your python code block must only contain the def
  solve_task(G): function definition and any necessary helper functions. Do NOT
  include any calls to the function outside of its definition.
  </python_api>
\end{alltt}

\subsection*{3D Scene Graph Python Interface}
The LLM is given a docstring style description of the \texttt{spark\_dsg} Python API which is shown below. The description covers the Python binding interface to the open source version of the library. Documentation for the library was generated with the assistance of Gemini 2.5 Pro \cite{comanici2025gemini} from the Pybind definitions and manually reviewed by a human expert. Example usage was added for functions which were less intuitive. For example, the node id field of \texttt{spark\_dsg} can be specified as a uint64 or a symbol containing a string and integer depending on the specific function used in the API. For our experiments we provided the documentation, the limited examples for specific functions, and a few complete examples of generically writing the \texttt{solve\_task(G)} function required for execution. This function expected the scene graph \texttt{G} as input and the model was instructed to solve the language query using the API and other common python libraries such as \texttt{numpy}.
\begin{alltt}
developer: # spark_dsg API Reference (Version 1.1.1)

Python interface (Dynamic Scene Graph) library, a hierarchical representation
for robotic perception. (import spark_dsg)
\end{alltt}

\begin{minted}{python}
## class DynamicSceneGraph:
"""The main container for a multi-layered, dynamic scene graph. It manages all
layers, nodes, and edges."""
def __init__(self, empty: bool, layer_keys: list[LayerKey], layer_names:
dict[LayerId, str]):
  """Creates a new DynamicSceneGraph (dsg). Can be initialized empty, or with
  a predefined set of layers.

  Args:
    empty (bool): If true, initializes an empty graph without default layers
    (optional, default=False).
    layer_keys (list[LayerKey]): A list of layer keys to pre-allocate.
    layer_names (dict[LayerId, str]): A map from layer ID to layer name.
  """
  # Example:
  # import spark_dsg as dsg
  # # Create a graph with default layers
  # graph = spark_dsg.DynamicSceneGraph()
  # # Create a completely empty graph
  # empty_graph = spark_dsg.DynamicSceneGraph(empty=True)

# Properties
nodes: iterator[SceneGraphNode]  # An iterator over all nodes in the graph
(including partitions).
unpartitioned_nodes: iterator[SceneGraphNode]  # An iterator over nodes in the
graph, excluding layer partitions.
edges: iterator[SceneGraphEdge]  # An iterator over all edges in the graph
(including partitions).
unpartitioned_edges: iterator[SceneGraphEdge]  # An iterator over edges in the
graph, excluding those connected to layer partitions.
layers: iterator[LayerView]  # An iterator over all non-partitioned layers in
the graph.
num_layers: int  # The total number of layers in the graph.
mesh: Mesh  # The 3D mesh associated with the scene graph.

# Methods
def has_layer(self, layer_id: int or str, partition_id: int) -> bool:
  """Checks if a layer exists in the graph.

  Args:
    layer_id (int or str): The integer ID or string name of the layer.
    partition_id (int): The partition ID (optional, default=0).
  Returns:
    bool: True if the layer exists, False otherwise.
  """
  # Example:
  # exists = graph.has_layer(spark_dsg.DsgLayers.OBJECTS)
def get_layer(self, layer_id: int or str, partition_id: int) -> LayerView:
  """Gets a read-only view of a specific layer.

  Args:
    layer_id (int or str): The integer ID or string name of the layer to retrieve.
    partition_id (int): The partition ID (optional, default=0).

  Returns: LayerView: A view of the requested layer. Throws an error if the
  layer does not exist.
  """
  # Example:
  # objects_layer = graph.get_layer(spark_dsg.DsgLayers.OBJECTS)
def has_node(self, node_id: NodeSymbol) -> bool:
  """Checks if a node exists anywhere in the graph.

  Args:
    node_id (NodeSymbol): The unique identifier of the node.
  Returns:
    bool: True if the node exists, False otherwise.
  """
def get_node(self, node_id: NodeSymbol) -> SceneGraphNode:
  """Gets a specific node from the graph.

  Args:
    node_id (NodeSymbol): The unique identifier of the node.
  Returns:
    SceneGraphNode: The requested node. Throws an error if the node does not exist.
  """
def find_node(self, node_id: NodeSymbol) -> SceneGraphNode or None:
  """Finds a specific node in the graph, returning None if not found.

  Args:
    node_id (NodeSymbol): The unique identifier of the node.
  Returns:
    SceneGraphNode or None: The requested node, or None if it does not exist.
  """
def has_edge(self, source_node: NodeSymbol, target_node: NodeSymbol) -> bool:
  """Checks if an edge exists between two nodes.

  Args:
    source_node (NodeSymbol): The ID of the source node.
    target_node (NodeSymbol): The ID of the target node.
  Returns:
    bool: True if the edge exists, False otherwise.
  """
def get_edge(self, source_node: NodeSymbol, target_node: NodeSymbol) -> SceneGraphEdge:
  """Gets the edge between two nodes.

  Args:
    source_node (NodeSymbol): The ID of the source node.
    target_node (NodeSymbol): The ID of the target node.

    SceneGraphEdge: The requested edge. Throws an error if the edge does not exist.
  """
def find_edge(self, source_node: NodeSymbol, target_node: NodeSymbol) ->
SceneGraphEdge or None:
  """Finds the edge between two nodes, returning None if not found.

  Args:
    source_node (NodeSymbol): The ID of the source node.
    target_node (NodeSymbol): The ID of the target node.
  Returns:
    SceneGraphEdge or None: The requested edge, or None if it does not exist.
  """
def get_position(self, node_id: NodeSymbol) -> numpy.ndarray[3]:
  """Gets the 3D position of a node.

  Args:
    node_id (NodeSymbol): The ID of the node.
  Returns:
    numpy.ndarray[3]: The (x, y, z) position of the node.
  """
def num_nodes(self, include_partitions: bool) -> int:
  """Gets the total number of nodes in the graph.

  Args:
    include_partitions (bool): Whether to include nodes in partitioned layers
    (optional, default=True).
  Returns:
    int:
  """
def num_edges(self, include_partitions: bool) -> int:
  """Gets the total number of edges in the graph.

  Args:
    include_partitions (bool): Whether to include edges connected to partitioned
    layers (optional, default=True).
  Returns:
    int:
  """
def empty(self) -> bool:
  """Checks if the graph has any nodes.

  Returns:
    bool:
  """
def clear(self):
  """Removes all nodes, edges, and layers from the graph.
  """
def clone(self) -> DynamicSceneGraph:
  """Creates a deep copy of the scene graph.

  Returns:
    DynamicSceneGraph:
  """
----------------------------------------
## class LayerView:
"""A read-only view of a single layer within the DynamicSceneGraph."""

# Properties
id: int  # The unique integer identifier for this layer.
partition: int  # The partition identifier for this layer.
nodes: iterator[SceneGraphNode]  # An iterator over all nodes in this layer.
edges: iterator[SceneGraphEdge]  # An iterator over all intra-layer edges in this layer.

# Methods
def num_nodes(self) -> int:
  """Gets the number of nodes in this layer.

  Returns:
    int:
  """
def num_edges(self) -> int:
  """Gets the number of intra-layer edges in this layer.

  Returns:
    int:
  """
def has_node(self, node_id: NodeSymbol) -> bool:
  """Checks if a node exists in this layer.

  Args:
    node_id (NodeSymbol): The ID of the node.
  Returns:
    bool: True if the node exists, False otherwise.
  """
def get_node(self, node_id: NodeSymbol) -> SceneGraphNode:
  """Gets a node from this layer.

  Args:
    node_id (NodeSymbol): The ID of the node.
  Returns:
    SceneGraphNode: The requested node. Throws an error if not found.
  """
def has_edge(self, source_node: NodeSymbol, target_node: NodeSymbol) -> bool:
  """Checks if an edge exists in this layer.

  Args:
    source_node (NodeSymbol): The ID of the source node.
    target_node (NodeSymbol): The ID of the target node.
  Returns:
    bool: True if the edge exists, False otherwise.
  """
def get_edge(self, source_node: NodeSymbol, target_node: NodeSymbol) -> SceneGraphEdge:
  """Gets an edge from this layer.

  Args:
    source_node (NodeSymbol): The ID of the source node.
    target_node (NodeSymbol): The ID of the target node.
  Returns:
    SceneGraphEdge: The requested edge. Throws an error if not found.
  """
def get_position(self, node_id: NodeSymbol) -> numpy.ndarray[3]:
  """Gets the 3D position of a node in this layer.

  Args:
    node_id (NodeSymbol): The ID of the node.
  Returns:
    numpy.ndarray[3]: The (x, y, z) position of the node.
  """
----------------------------------------
## class SceneGraphNode:
"""Represents a single node in the scene graph."""

# Properties
id: NodeSymbol  # The unique identifier for the node.
layer: int  # The ID of the layer this node belongs to.
attributes: NodeAttributes  # A container for the node's
attributes (e.g., position). The actual type may be a subclass
like SemanticNodeAttributes.

# Methods
def has_parent(self) -> bool:
  """Checks if the node has a parent.
  Returns:
    bool:
  """
def has_siblings(self) -> bool:
  """Checks if the node has siblings.
  Returns:
    bool:
  """
def has_children(self) -> bool:
  """Checks if the node has children.
  Returns:
    bool:
  """
def get_parent(self) -> NodeSymbol or None:
  """Gets the parent of the node, if one exists.
  Returns:
    NodeSymbol or None:
  """
def siblings(self) -> list[NodeSymbol]:
  """Gets a list of the node's siblings.
  Returns:
    list[NodeSymbol]:
  """
def children(self) -> list[NodeSymbol]:
  """Gets a list of the node's children.
  Returns:
    list[NodeSymbol]:
  """
def parents(self) -> list[NodeSymbol]:
  """Gets a list of the node's parents (can be more than one).
  Returns:
    list[NodeSymbol]:
  """
def connections(self) -> list[NodeSymbol]:
  """Gets a list of all nodes connected to this one (parents and children).
  Returns:
    list[NodeSymbol]:
  """
----------------------------------------
## class SceneGraphEdge:
"""Represents a directed edge between two nodes."""

# Properties
source: NodeSymbol  # The ID of the node where the edge originates.
target: NodeSymbol  # The ID of the node where the edge terminates.
info: EdgeAttributes  # A container for the edge's attributes (e.g., weight).
----------------------------------------
## class NodeSymbol:
"""A unique identifier for a node, composed of a character prefix and a numerical ID.
E.g., 'p1' for place 1. (Case Sensitive)"""
def __init__(self, key: char, index: int, value: int):
  """Creates a new NodeSymbol.
  Args:
    key (char): The character prefix for the category (e.g., 'p').
    index (int): The numerical index for the node.
    value (int): A raw integer value to construct the symbol from.
  """
  # Example:
  # node_id = spark_dsg.NodeSymbol('p', 10)
  # node_id_from_val = spark_dsg.NodeSymbol(node_id.value)

# Properties
category: str  # The character prefix of the symbol (e.g., 'p').
category_id: int  # The numerical part of the symbol (e.g., 10).
value: int  # The full integer representation of the symbol.

# Methods
def str(self, literal: bool) -> str:
  """Gets the string representation of the symbol.
  Args:
    literal (bool): If true, show category name instead of char (optional, default=True).
  Returns:
    str: The string label (e.g., 'p10').
  """
  # Example:
  # symbol = spark_dsg.NodeSymbol('p', 10)
  # label = symbol.str() # 'p10'
----------------------------------------
## class NodeAttributes:
"""The base class for attributes associated with a node."""
def __init__(self):
  """Creates default node attributes.
  """
# Properties
position: numpy.ndarray[3]  # The 3D position (x, y, z) of the node.
is_active: bool  # Whether the node is considered part of the active window.
is_predicted: bool  # Whether the node is a predicted entity.
last_update_time_ns: int  # Timestamp of the last update in nanoseconds.
----------------------------------------
## class SemanticNodeAttributes:
"""Attributes for nodes with semantic meaning. Inherits from `NodeAttributes`."""
def __init__(self):
  """Creates default semantic node attributes.
  """

# Properties
name: str  # A human-readable name for the node (e.g., 'object_1').
color: numpy.ndarray[3]  # The RGB color of the node as a uint8 array.
bounding_box: BoundingBox  # The bounding box of the node.
semantic_label: int  # The semantic label ID for the node
(e.g., from a segmentation model).
semantic_feature: numpy.ndarray  # An arbitrary semantic feature vector for the node.
----------------------------------------
## class ObjectNodeAttributes:
"""Attributes for object nodes. Inherits from `SemanticNodeAttributes`."""
def __init__(self):
  """Creates default object node attributes.
  """

# Properties
registered: bool  # Whether the object has been registered against a database.
world_R_object: Quaternion  # The orientation of the object in the world frame.
mesh_connections: list[int]  # List of vertex indices from the scene mesh that
belong to this object.
----------------------------------------
## class EdgeAttributes:
"""A container for attributes associated with an edge."""
def __init__(self):
  """Creates default edge attributes.
  """

# Properties
weighted: bool  # Indicates if the edge has a meaningful weight.
weight: float  # The weight of the edge, often used for pathfinding.
----------------------------------------
## class BoundingBox:
"""Represents a 3D bounding box, which can be axis-aligned (AABB)
or oriented (OBB)."""
def __init__(self):
  """Creates a new BoundingBox. Multiple constructors are available for
  different box types.
  """
  # Example:
  # import numpy as np
  # # Create an AABB from min and max corners
  # box = spark_dsg.BoundingBox(np.array([-1, -1, -1], dtype=np.float32),
      np.array([1, 1, 1], dtype=np.float32))

# Properties
min: numpy.ndarray[3]  # The minimum corner of the box in the world frame.
max: numpy.ndarray[3]  # The maximum corner of the box in the world frame.
type: BoundingBoxType  # The type of bounding box (e.g., AABB, OBB).
dimensions: numpy.ndarray[3]  # The (x, y, z) dimensions of the box.
world_P_center: numpy.ndarray[3]
  # The position of the box's center in the world frame.
world_R_center: numpy.ndarray[3,3]
  # The rotation matrix of the box's center in the world frame.

# Methods
def is_valid(self) -> bool:
  """Checks if the bounding box is valid.
  Returns:
    bool:
  """
def volume(self) -> float:
  """Computes the volume of the bounding box.
  Returns:
    float:
  """
def corners(self) -> list[numpy.ndarray[3]]:
  """Gets the 8 corners of the bounding box in the world frame.
  Returns:
    list[numpy.ndarray[3]]:
  """
def intersects(self, other: BoundingBox) -> bool:
  """Checks if this bounding box intersects with another.

  Args:
    other (BoundingBox): The other bounding box to check against.
  Returns:
    bool:
  """

# Enums
class BoundingBoxType:
  INVALID = ...
  AABB = ...
  OBB = ...
  RAABB = ...
----------------------------------------
## class Quaternion:
"""A representation of a rotation in 3D space."""
def __init__(self, w: float, x: float, y: float, z: float):
  """Creates a new Quaternion.
  Args:
    w (float): The real part.
    x (float): The first imaginary part.
    y (float): The second imaginary part.
    z (float): The third imaginary part.
  """
  # Example:
  # # Identity rotation
  # q = spark_dsg.Quaternion(1.0, 0.0, 0.0, 0.0)

# Properties
w: float  # The w (real) component of the quaternion.
x: float  # The x component of the quaternion.
y: float  # The y component of the quaternion.
z: float  # The z component of the quaternion.
----------------------------------------
class DsgLayers:
"""Standard integer identifiers for common semantic layers in the scene graph."""
SEGMENTS = ...  # Layer for mesh segments or supervoxels.
OBJECTS = ...  # Layer for individual objects.
AGENTS = ...  # Layer for dynamic agents, like robots or people.
PLACES = ...  # Layer for places (3D)
MESH_PLACES = ...  # Layer for places defined by a collection of mesh vertices.
  (2D Places)
ROOMS = ...  # Layer for rooms.
BUILDINGS = ...  # Layer for entire buildings or structures.
\end{minted}

\subsection*{In-Context Scene Graph Method In-Context Examples}

\begin{alltt}
Here are some examples of mapping from natural language instructions to PDDL goals 
paired with robot IDs. Each example uses the 3D scene graph below. Note that this
3D scene graph is just for these examples and you should NOT use this when 
answering the new instruction.
<Example 3D Scene Graph>
Objects:
- (id=O0, type=tree, pos=(-3.14,1.13,0.1), parent_places={'p4'})
- (id=O1, type=vehicle, pos=(3.34,3.53,0.1), parent_places={'p0'})
- (id=O2, type=door, pos=(3.33,3.48,0.2), parent_places={'p4'})
- (id=O3, type=tree, pos=(4.47,-4.72,-0.1), parent_places={'p5'})
- (id=O4, type=vehicle, pos=(-2.51,6.63, 0.2), parent_places={'p3'})
- (id=O5, type=boat, pos=(1.34,3.28,-0.2), parent_places={'p2'})
- (id=O6, type=seating, pos=(1.37,3.03,0.01), parent_places={'p2'})
- (id=O7, type=tree, pos=(9.10,-2.01,0.04), parent_places={'p6'})
Places:
- (id=p0, siblings={'p1','p4'}, parent_rooms={'R0'})
- (id=p1, siblings={'p0','p4'}, parent_rooms={'R0'})
- (id=p2, siblings={'p3'}, parent_rooms={'R1'})
- (id=p3, siblings={'p2'}, parent_rooms={'R1'})
- (id=p4, siblings={'p0','p1'}, parent_rooms={'R0'})
- (id=p5, siblings={'p6'}, parent_rooms={'R2'})
- (id=p6, siblings={'p5'}, parent_rooms={'R2'})
Rooms:
- (id=R0, type=parking_lot, pos=(1.17,2.71,0.01), siblings=none)
- (id=R1, type=dock, pos=(0.67,4.31,0.09), siblings=none)
- (id=R2, type=courtyard, pos=(6.79,2.31,-0.01), siblings=none)
</Example 3D Scene Graph>
"

\textbf{user}: "Go to object 1."
\textbf{assistant}: "(visited-object O1)"
\textbf{user}: "Move to o1."
\textbf{assistant}: "(visited-object O1)"
\textbf{user}: "Head to place 100."
\textbf{assistant}: "(visited-place P100)"
\textbf{user}: "Head to room 21."
\textbf{assistant}: "(visited-place R21)"
\textbf{user}: "Drive over to the door and visit the boat."
\textbf{assistant}: "(and (visited-object O2) (visited-object O5))"
\textbf{user}: "Walk over to the door or end up by the boat."
\textbf{assistant}: "(or (visited-object O2) (visited-object O5))"
\textbf{user}: "Head over to the door and move to the boat."
\textbf{assistant}: "(and (visited-object O2) (visited-object O5))"
\textbf{user}: "Could you navigate to the vehicle on the dock?"
\textbf{assistant}: "(visited-object O4)"
\textbf{user}: "Head to the rightmost vehicle."
\textbf{assistant}: "(visited-object O1)"
\textbf{user}: "Move to the vehicle on the left."
\textbf{assistant}: "(visited-object O4)"
\textbf{user}: "I need you to go to the parking lot and the dock."
\textbf{assistant}: "(and (visited-place R0) (visited-place R1))"

\end{alltt}

\subsection*{SLDP Answer Format Explanation}
For the question and answer task, the following description expected answer formatting is used instead of requesting a PDDL goal:
\begin{alltt}
### Syntax
A primitive string is a sequence of alphanumeric characters (with no quotation).
A primitive number is a floating point representation of a number.
A `list` is written as `[element1, element2, ... elementN]`
A `set` is written as `<element1, element2, ... elementN>`
A `dict` is written as `\{k1: v1, k2: v2\}`
A `point` is written as `POINT(x y z)` (note the lack of comma)
### Denoting Final Answer:
Format your final answer (*not* any intermediate tool calls) as an SLDP
expression wrapped between <answer> and </answer> tags, such as
<answer><1,2,3></answer>. Only a single pair of answer tags should appear in
your solution.
Your answer should be an SLDP set
\end{alltt}
The final line is updated to request the answer type (number, string, list, set, dict, point) for each question.
\end{document}